\documentclass[10pt,conference]{IEEEtran}
\IEEEoverridecommandlockouts
\usepackage{cite}
\usepackage{amsmath,amssymb,amsfonts}
\usepackage{algorithmic}
\usepackage{graphicx}
\usepackage{textcomp}
\usepackage{xcolor}
\usepackage{epsfig}
\usepackage{booktabs}
\usepackage{CJKutf8}        
\usepackage{bbding}
\usepackage{algorithm}
\usepackage{algorithmic}
\usepackage{float}
\usepackage{authblk}
\usepackage[colorlinks,linkcolor=black,citecolor=black]{hyperref}
\def\BibTeX{{\rm B\kern-.05em{\sc i\kern-.025em b}\kern-.08em
    T\kern-.1667em\lower.7ex\hbox{E}\kern-.125emX}}
\begin{document}


\title{SARAD: LLM-Based Safety-Aware Hybrid Reinforcement Learning with Collision Prediction for Autonomous Driving\\
\thanks{This work is supported by National Natural Science Foundation (NNSF) of China under Grant 62373100 and 62233003, and National Science and Major Project under Grant 2021ZD0112702.}

}


\author{
	Kangyu Wu$^{1,2,*}$,
	Peng Cui$^{1,2,*}$,
	Guoxi Chen$^{1,2}$,
	Ya Zhang$^{1,2,\dagger}$
}
\affil{
	$^{1}$School of Automation, Southeast University, Nanjing, China\\
	$^{2}$Key Laboratory of Measurement and Control of Complex Systems of Engineering,\\ Ministry of Education,  Nanjing, China\\
	$^{*}$These authors contributed equally to this work.\\
	$^{\dagger}$Corresponding Authors: Ya Zhang (yazhang@seu.edu.cn)
}


\setlength{\affilsep}{3pt}	

\maketitle

\begin{abstract}
Ensuring both safety and efficiency in decision-making of autonomous driving systems remains a fundamental challenge. Traditional Deep Reinforcement Learning (DRL) suffers from unsafe random exploration and slow convergence, while Large Language Models (LLMs) demonstrate inherent latency in real-time inference operations. To address these limitations, this paper proposes SARAD, a novel safety-aware hybrid framework that synergizes LLMs and DRL for autonomous driving. SARAD substitutes DRL's random exploration with Retrieval-Augmented Generation (RAG)-enhanced, LLM-guided decisions sourced from a dynamic expert knowledge repository. An attention discriminator that integrates LLM’s prior knowledge into DRL policy optimization is proposed. A collision predictor module, which is fine-tuned with historical collision data, is further designed to guarantee the safety of the vehicle. Extensive experiments show that SARAD achieves significant performance improvements in Highway-Env simulator, validating the effectiveness of the model in autonomous driving.
\end{abstract}

\begin{IEEEkeywords}
Autonomous Driving, Reinforcement Learning, Large Language Model, Retrieval-augmented Generation
\end{IEEEkeywords}

\section{Introduction}
In autonomous driving, decision-making and control algorithms are essential for vehicle operation. Traditional decision-making algorithms include Finite State Machines and tree-based models. The advent of Reinforcement Learning (RL) has introduced significant potential for the decision-making process in autonomous driving \cite{kiran2021deep}. Due to the capability for autonomous optimization within complex dynamic environments, DRL has been established as a crucial methodology for training driving policies. However, traditional DRL relies on a large amount of random exploration to accumulate experience, which is particularly dangerous in highway scenarios. On the one hand, random exploration may cause vehicles to frequently enter high-risk states (e.g., sharp braking, lane-change conflicts), which increases the safety hazards in real-world testing. On the other hand, inefficient random exploration which is particularly detrimental in high-speed dynamic scenarios substantially prolongs policy convergence.
\begin{figure}[!t]
	\centering
	{\epsfig{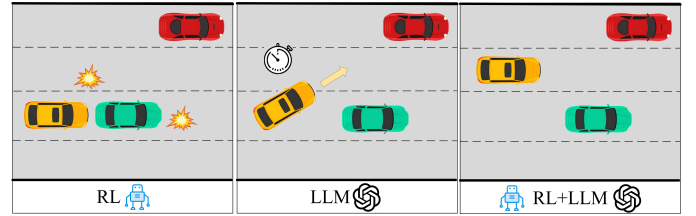}}
	\caption{A comparison of various autonomous driving decision-making methods. It reveals that using Reinforcement Learning alone can lead to behavioral uncertainty, while relying solely on Large Language Models may introduce unacceptable reasoning latency. To address these limitations, this paper proposes a novel approach that combines the complementary strengths of DRL and LLMs.}
	\label{fig:1}
\end{figure}

Large Language Models have played an irreplaceable role across diverse domains. While approaches like DiLu demonstrate promising results in linguistically structuring traffic scenarios \cite{wen2023dilu}, leveraging LLMs (e.g., GPT \cite{brown2020language}) for vehicular decision-making \cite{xu2023drivegpt4}, or fine-tuning LLMs with driving-style datasets, they exhibit critical limitations. Notably, their lack of real-time responsiveness hinders the immediate generation of optimal driving actions, thereby impeding practical deployment. Consequently, it remains challenging to construct autonomous driving decision systems that rely solely on LLMs.

To address these challenges, this paper proposes SARAD, a novel framework leveraging LLMs to assist DRL training. As shown in Fig. \ref{fig:1}, SARAD replaces traditional random exploration with LLM-guided decisions, synergistically combining the inference efficiency of DRL with the superior reasoning of LLMs. Concurrently, a discriminator-guided policy optimization mechanism is designed to effectively incorporate the action guidance derived from the LLM's prior knowledge into the DRL agent's learning process. Therefore, we construct a dynamic knowledge repository of expert driving experience for highways. This knowledge repository is continuously updated during vehicle operation with real-time expert insights and critical scenarios (e.g., collision avoidance). Furthermore, Retrieval-Augmented Generation (RAG) extracts representative driving experiences from this repository \cite{lewis2020retrieval}, enabling LLMs to generate safer, human-aligned actions through contextual learning. This approach significantly reduces unproductive exploration during policy optimization. While absolute safety remains the fundamental objective of autonomous driving, strategies generated by LLMs or DRL may still fail to prevent collisions in edge cases. Such limitations arise from incomplete knowledge coverage or reasoning deficiencies in decision-making systems. To address this critical vulnerability, we propose a final safety layer featuring a collision prediction module. This mechanism continuously assesses maneuver risk levels and activates fallback protocols when imminent danger is detected. By serving as a proactive safety safeguard, this system mitigates collision risks even when primary decision strategies fail.
	
Overall, the main contributions of this paper are as follows: 

(1) A novel decision-making paradigm synergistically combining LLMs and DRL is proposed. It replaces stochastic exploration with LLM guidance, significantly accelerating DRL convergence, enhancing final cumulative rewards, and ensuring inference efficiency.

(2)  A learning strategy based on discriminators is designed to effectively integrate the prior actions of LLM into the optimization process of DRL.

(3) A collision risk prediction module, which is developed based on a fine-tuned LLM and a set of rule-based safety strategies, is designed to assess the next collision risk and mitigate it promptly and effectively.
	
The source code supporting this study will be made available following the final code refactoring and documentation updates.

\begin{figure*}[t]
\centering
{\epsfig{file = 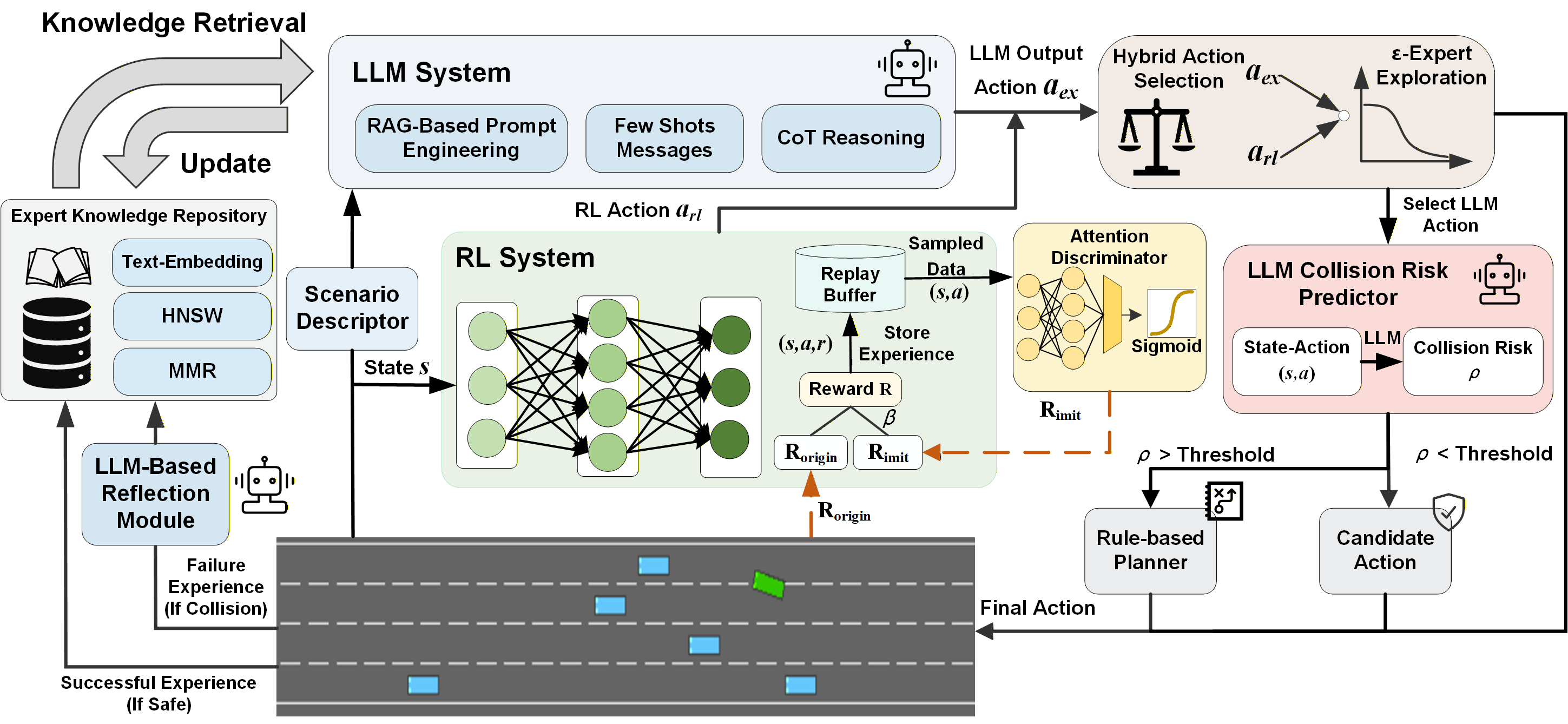, width = \textwidth}}
\caption{Overview of the proposed framework. This framework includes two systems, including DRL and LLM, where LLM assists DRL for training. Within the description of current state, the RAG-based LLM outputs a reasonable decision action. After the evaluation of the collision predictor, the rule-based strategy for safety will be used once the risk is too high, and the LLM will be used for reflection. The planned actions will be combined with DRL through the discriminator module to facilitate DRL to learn the decision-making strategy of the LLM.}
	\label{fig:2}
\end{figure*}

\section{Related Work}
\subsection{DRL-Based Decision-making for Autonomous vehicles}
Deep Reinforcement Learning is pivotal in autonomous driving due to its adaptive learning in complex scenes. Early foundations like DQN \cite{mnih2015human} and Double DQN \cite{van2016deep} addressed discrete decision-making and Q-value overestimation. For continuous control, DDPG \cite{lillicrap2015continuous} and SAC \cite{tang2022highway} integrated Actor-Critic frameworks with entropy regularization to enhance exploration and smoothness. To handle rare scenarios, meta-learning approaches like MAML \cite{finn2017model} enable rapid policy adaptation. In imitation learning, GAIL \cite{ho2016generative} and IRL \cite{kuderer2015learning} bypass reward design by mimicking human behavior.


\subsection{LLM for Autonomous Driving}
Large Language Models are being integrated into driving via two main paths: fine-tuning and prompt engineering. 

Fine-tuning adapts LLMs to multi-modal driving data. Frameworks like DriveGPT4 \cite{xu2023drivegpt4} fuse BEV features or high-resolution images for deeper scene understanding. DriveLM \cite{sima2024drivelm} discretizes control signals into tokens, while Agent-Driver \cite{mao2023language} utilizes modular toolchains and to bridge the perception-decision gap.

Prompt engineering leverages zero-shot capabilities. DiLu \cite{wen2023dilu} utilizes a RAG-based memory architecture with Chain of Thought (CoT) cues. The RRR framework \cite{cui2024receive} and LanguageMPC \cite{sha2023languagempc} establish logical hierarchies for tool invocation and controller tuning, while PlanAgent \cite{zheng2024planagent} drives decision-making via textualized environment descriptions.

Integrating LLMs with DRL enhances sample efficiency and safety, GLAM \cite{carta2023grounding} grounds LLMs via online RL, while ELLM \cite{du2023guiding} generates context-aware exploration goals. LGDRL \cite{pang2024large} employs LLM-guided policy constraints to reduce expert reliance. Furthermore, LLM-enhanced RLHF \cite{sun2024optimizing} aligns driving policies with human safety preferences.

\section{Methods}
This paper leverages LLMs to enhance the training of DRL, achieving faster convergence and higher reward. It is assumed that there is one autonomous vehicle and $N$ other vehicles in the highway environment. The decision-making problem of the autonomous vehicle can be modeled as a Markov Decision Process:
\begin{equation}
    MDP=(S,A,P,R,\gamma),
\end{equation}
where state space $S\in \mathbb{R}^{4*(N+1)}$ represents the autonomous vehicle's observation of the current environment. It is assumed that the observation space equals the state space, i.e., each observation fully captures the environment's state. At timestep $t$, the observed state $S_t$ encapsulates critical dynamic information of both the ego vehicle and surrounding vehicles, formally defined as:
\begin{equation}
    s_t=(x_{ego},y_{ego},v^x_{ego},v^y_{ego},\{x_i,y_i,v^x_i,v^y_i\}^{N}_{i=1}),
\end{equation}
where $x,y,v^x,v^y$ denote the coordinates and velocity in the x and y directions, respectively. In this paper, the ego vehicle's action space $A\in \mathbb{Z}$ is defined as a discrete action space, comprising five actions: accelerate, decelerate, change lanes to the left, change lanes to the right, and maintain the speed. $P(s_{t+1}|s_t,a_t)$ models state transitions under kinematic constraints and environmental stochasticity. $R$ and $\gamma$ denote the reward offered to the agent and the discount factor for the reward, respectively.

\subsection{Overall framework}
The proposed framework consists of three parts: the LLM-enhanced DRL Framework with an Attention Discriminator, the RAG and reflection module based on the expert knowledge repository, and the collision predictor. An overview of the framework is shown in Fig. \ref{fig:2}. In this paper, the classical DQN algorithm is adopted as the baseline for deep reinforcement learning and modified with LLM. Specifically, we replace the random exploration in the original DQN with LLM, design a discriminator to facilitate the integration of LLM's prior expert actions into the reinforcement learning policy, and simultaneously optimize both the reinforcement learning policy network and the discriminator. To strengthen LLMs' traffic scenario interpretation and decision execution, we develop an expert knowledge repository for driving expertise encapsulation. It introduces a RAG module to retrieve similar driving experiences based on available information before making decisions for the LLM. To enhance the robustness of the autonomous driving system, a reflection module is added to the system for describing and reflecting on the states and actions that lead to a collision when the agent experiences one, and for incorporating the results of the reflection into the expert knowledge repository. Finally, an LLM-based collision predictor is designed. The system enables the underpinning strategy to maximize the agent's safety when the collision risk is too high.

\subsection{LLM-enhanced DRL Framework with an Attention Discriminator}
The $\epsilon$-greedy strategy is often used to control the frequency of random exploration, gradually decreasing it over time. This phase exhibits significant inefficiency and computational latency, prolonging algorithm convergence and amplifying early-stage uncertainty. In order to alleviate the disadvantages caused by the random exploration in the early stage in traditional reinforcement learning algorithms, this paper proposes a new framework that substitutes random exploration with LLM-guided decisions, transforming the traditional $\epsilon$-greedy strategy into an $\epsilon$-expert exploration mechanism, which progressively reduces LLM utilization during later training stages.

To address the distribution mismatch introduced by LLM-guided exploration while avoiding direct behavior cloning, we introduce an attention-based discriminator $\mathcal{D}$ as a reward-shaping mechanism rather than a pure imitation module. Unlike conventional imitation learning frameworks that explicitly force the agent to replicate expert actions, the proposed discriminator learns to measure semantic consistency between the agent’s actions and LLM-guided decisions, and provides a continuous imitation reward to softly regularize the policy optimization process. The discriminator serves as a bridge between high-level LLM guidance and low-level policy learning, encouraging the agent to stay close to LLM-recommended behaviors during early training while still allowing deviations when environmental rewards dominate. This design alleviates the generalization errors commonly observed in pure behavior cloning and enables a smooth transition from LLM-guided exploration to autonomous policy optimization. 

Specifically, the state input $s_t$ is processed through convolutional layers and a Squeeze-and-Excitation (SE) block to output channel-wise attention weights $\mathbf{s}$:
\begin{equation}
	\mathbf{s} = \sigma(\mathbf{W}_2 \delta(\mathbf{W}_1 \mathbf{z})),
\end{equation}
where $\mathbf{z}$ is the squeezed channel descriptor, $\mathbf{W}_1$ and $\mathbf{W}_2$ are learnable weight matrices, and $\delta$ and $\sigma$ denote the ReLU and Sigmoid activation functions, respectively. The recalibrated state features are then flattened, concatenated with the one-hot encoded discrete action $a_t$, and passed through fully connected layers with a Sigmoid activation to compute the probability $\mathcal{D}(s_t,a_t)$. Finally, the imitation reward $R_{imit}$ is added to the original reward $R_{origin}$, with the calculation formula as follows:
\begin{equation}
    R_{imit}=-log(1-\mathcal{D}(s_t,a_t)+\eta),
    \label{equ:3}
\end{equation}
\begin{equation}
    R_{origin} = a\frac{{v - {v_{\min }}}}{{{v_{\max }} - {v_{\min }}}} - b \cdot \mathbb{I}_{\text{col}},
\end{equation}
\begin{equation}
    R_{total}=R_{origin}+\beta \cdot R_{imit},
\end{equation}
where $R_{total}$ is the final reward, $s_t$, $a_t$ and $v$ represent the state, action and speed of the ego vehicle, respectively, $v_{min}$ and $v_{max}$ are the minimum and maximum speeds during driving. $\mathbb{I}_{\text{col}} \in \{0,1\}$ is a collision indicator, where $\mathbb{I}_{\text{col}} = 1$ indicates a collision. $\eta$, $a$, $b$ and $\beta$ are hyperparameters. In this process, both the discriminator and the agent's policy network are trained at the same time. The loss functions for both are shown below:
\begin{equation}
    \mathcal{L}_D=-\frac{1}{B}\sum^{B}_{i=1}[log\mathcal{D}(s^i_{ex},a^i_{ex})+log(1-\mathcal{D}(s^i_{ag},a^i_{ag}))],
\end{equation}
\begin{equation}
    \mathcal{L}_{policy}=\frac{1}{B}\sum^{B}_{i=1}\mathcal{L}(Q(s^i_t,a^i_t),R^i_{total}+\gamma\underset{a}{\max}Q_{target}(s^i_{t+1},a)),
\end{equation}
where Huber loss is used for $\mathcal{L}$, $\mathcal{L}_D$ and $\mathcal{L}_{policy}$ are used to train the discriminator $\mathcal{D}$ and the DQN network, respectively. $B$ presents batch size, $i$ denotes the $i$-th sample within a batch, and the subscripts $ex$ and $ag$ denote expert and agent, respectively. $Q$ and $Q_{target}$ represent two networks in DQN used to calculate Q value and target Q Value.

It is crucial to note the theoretical distinction from traditional Generative Adversarial Imitation Learning. While sharing a similar adversarial loss, our discriminator evaluates the "semantic consistency" by employing the SE attention module to map the agent's spatial maneuvers against the LLM's text-derived logical priors, acting as a soft regularization term rather than forcing strict distribution matching. 

During LLM-based decision-making, the current state observation $s_t$ is first obtained. A textual description of the scene is then generated through a prompt generator, incorporating details such as vehicle position, speed, decision-making goals, and other essential information. Subsequently, the prompt is enriched using an RAG module, with detailed implementation described in the following subsection.

\begin{figure*}[t]
	\centering
	{\epsfig{file = 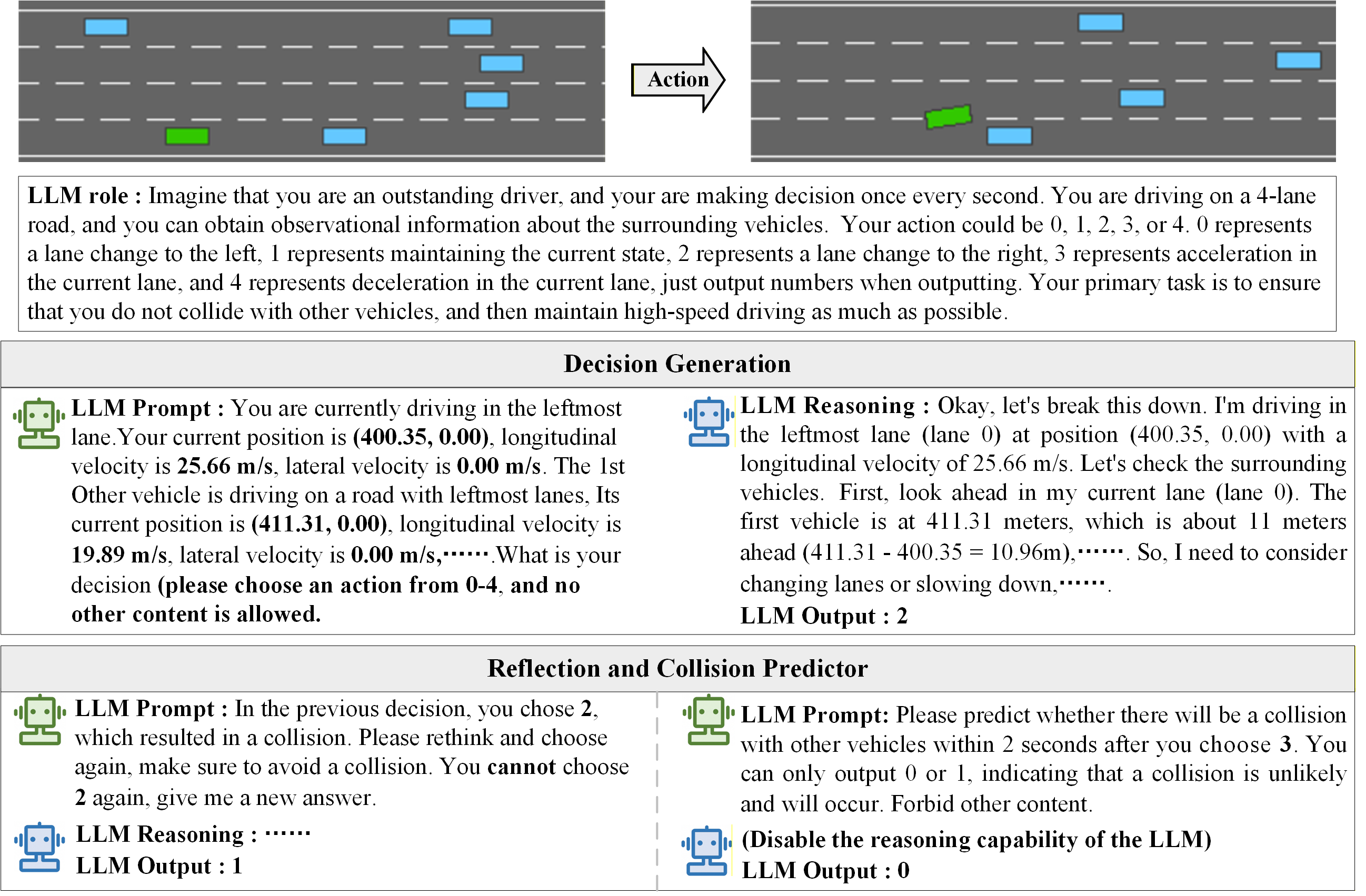, width = \textwidth}}
	\caption{Generated scene description and Prompt construction. LLM only outputs the action of the decision after reasoning.}
	\label{fig:3}
\end{figure*}

\subsection{RAG and Reflection Module}
Since pre-trained LLMs face challenges in developing specific understandings of the autonomous driving environment addressed in this work, sole reliance on them yields unreliable tactical decisions. To address this problem, this paper introduces a RAG mechanism to enhance the efficiency of experience utilization. This mechanism is implemented by constructing a driving experience vector database, which enhances the retrieval of similar experiences and ultimately facilitates interaction with the LLM for collaborative decision-making. Specifically, the RAG mechanism is designed to bridge long-term memory and real-time perception, enabling the system to retrieve and reuse prior knowledge instead of relying solely on the LLM’s limited context window.

When no collision occurs, the agent structurally encodes state, observations, decision actions, and environmental feedback generated during driving with a 20\% probability. These encoded data are stored as low-dimensional, dense vectors in a vector database, forming a knowledge repository that covers multiple driving scenarios. The vector database is continuously updated during driving, ensuring that new scenarios, once encoded, are available for subsequent retrieval. Before each decision, the system generates a context-aware embedding representation of the current observation through the Embedding model. It retrieves the top $k$ historical experience vectors with the highest semantic relevance to the current scene from the database. These retrieved vectors are then concatenated with the current observation and input into the LLM. In practice, this concatenation process is implemented by designing a retrieval-augmented prompt, in which the retrieved experiences are reformulated into structured natural language descriptions and appended to the current query. Prompt engineering is employed to guide the model in fusing historical priors with real-time states, thereby generating decision suggestions.

To further improve decision interpretability and gradually enhance the LLM’s decision quality, this paper also designs a collision-driven reflection module. When the agent collides, the system reconstructs the reflection prompt, adds the historical decision actions that led to the collision, makes the LLM decide again, generates an alternative risk-avoidance strategy, and appends it to the vector database. This reflection prompt is designed in a chain-of-thought style, which explicitly forces the LLM to analyze the cause of failure before proposing a new strategy, thus improving both safety and reasoning transparency. This reflection module significantly enhances the robustness of the proposed autonomous driving system in long-tail scenarios. The decision-making and reflection process for LLM is depicted in Fig. \ref{fig:3}.

\subsection{Collision Predictor}
This paper designs a collision predictor to enhance the robustness of the autonomous driving system. By computing real-time collision probabilities for potential maneuvers and implementing rule-based fallback strategies, this module mitigates potential accidents. The predictor operates as an independent safety layer parallel to the LLM decision module, continuously monitoring candidate actions before execution and estimating their collision likelihood based on learned risk patterns. Specifically, when LLM-generated decisions are predicted to yield high collision probabilities, the agent activates predefined safety rules, thereby providing fail-safe protection against potentially hazardous LLM outputs. These safety rules include emergency braking, lane-keeping, and maintaining a minimum safe distance, which are triggered according to different levels of predicted risk. 

To implement the collision predictor, we fine-tuned a pre-trained 0.6B-parameter LLM with the objective of enhancing its awareness of safety-critical patterns and collision risks in autonomous driving scenarios. The fine-tuning objective was formulated as a binary collision risk prediction task, mapping state-action representations to a collision probability score $\in [0, 1]$, which is then compared against a pre-defined threshold to determine whether to override the LLM's decision. The training data set was collected in advance through a simulation environment, where negative samples were strictly selected from the state-action pairs of the three frames preceding the collision in order to capture the critical risk features. In comparison, the positive samples were randomly selected from moments without collision risk for three frames, ensuring that each of the positive and negative samples contained 20,000 samples. The balanced dataset design ensures that the model learns to discriminate not only between safe and unsafe scenarios but also to generalize across diverse driving conditions. All the samples were shuffled and used for the fine-tuning of the LLM.

\section{Experiment}
\subsection{Experimental Setup}
We utilize the highway-env simulator to construct the experimental scenario \cite{leurent2018environment}. The experimental scenario consists of four driving lanes, each 4 meters in width and modeled as infinitely long. A total of 10 moving vehicles are present in the scenario, including one ego vehicle. The action space of each vehicle is discrete, with only five permissible actions: changing lane to the left, changing lane to the right, accelerating straight, decelerating straight, and maintaining speed. We assume that the autonomous vehicle can access complete information about itself and all other vehicles, including coordinates and velocities in both x and y directions. The experiment is conducted over 160000 episodes, where an episode is defined as terminated if a collision occurs between the AV and other vehicles or if the driving time exceeds 40 seconds. We used three metrics to evaluate the experimental results: average runtime of each episode, average reward of each episode, and total reward of each episode, the metrics are defined as follows:
\begin{equation}
    Avg Runtime=t_{n}-t_{0},
\end{equation}
\begin{equation}
    Avg Reward=\frac{\sum_{i=0}^{n}reward_i}{n},   
\end{equation}
\begin{equation}
    Total Reward=\sum_{i=0}^{n}reward_i,
\end{equation}
where $n$ indicates the total number of steps in an episode, $t_i$ represents the timestamp at the $i^{th}$ step, and $reward_i$ represents the reward at the $i^{th}$ step.

The baseline algorithm utilizes the Deep Q-Network (DQN), a classic DRL algorithm designed for discrete action spaces. As previously described, this algorithm features an action space size of 5, a learning rate of $1 \times 10^{-4}$, a batch size of 32, a replay buffer capacity of 15000, and a discount factor $\gamma$ of 0.9, with its exploration rate ultimately converging to 0.05. In the reward function, $a$ and $b$ are set to 0.4 and 1, respectively. $collision=1$ if and only if a collision occurs, otherwise, $collision=0$, and $\beta$, the weight of imitation reward, is set to 0.2, $\eta$ is set to $10^{-8}$. The observed state space is structured using the 2D coordinates and velocities of 10 vehicles.

In order to trade off the inference effect and speed of the LLM in action decision-making, Qwen-Plus is selected \cite{team2024qwen2}, and the Chroma database and Text-embedding-v2 are used to build an expert knowledge repository. Meanwhile, the RAG mechanism is used to enhance the LLM, while Hierarchical Navigable Small World (HNSW) \cite{xiao2024enhancing} with Maximal Marginal Relevance (MMR) \cite{guo2010probabilistic} is applied in a hybrid retrieval strategy. For the LLM in the collision predictor, Qwen3-0.6B is selected to ensure absolute inference speed \cite{yang2025qwen3}. LoRA (configured with a rank of 8, a learning rate of $5 \times 10^{-5}$ using a cosine scheduler, 5 epochs, and FP16 precision) and customized datasets are also utilized for fine-tuning, which ensures a certain degree of generalization and accuracy \cite{hu2022lora}. To validate methodological generalizability, a machine learning predictor using Gradient Boosting Decision Trees (GBDT) with Logistic Regression (LR) serves as a control baseline, employing feature engineering to extract spatial coordinates and velocity vectors from state representations for binary classification training \cite{xu2022gbdt}.

\begin{figure*}[t]
	\centering
	{\epsfig{file = 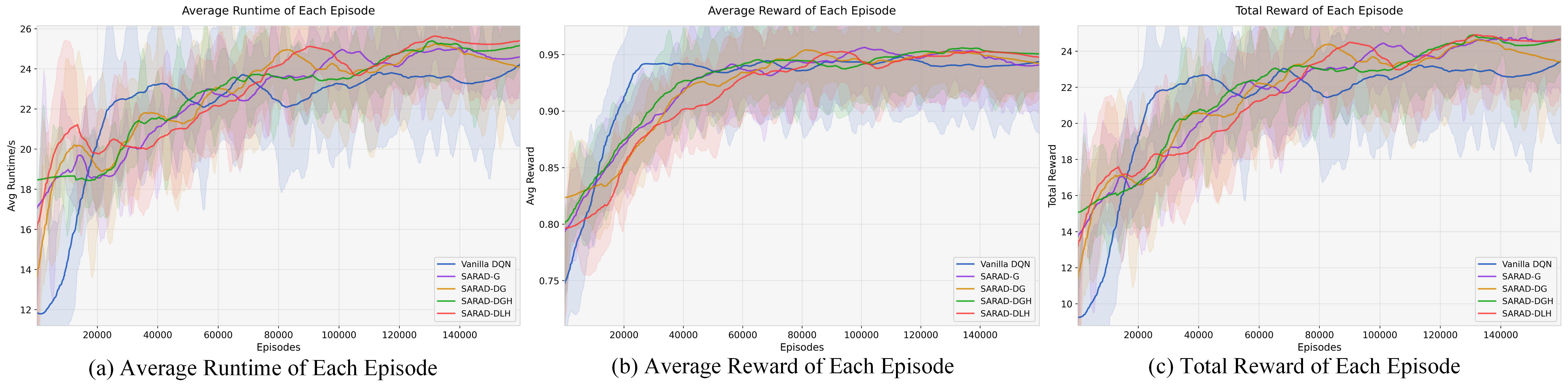, width = \textwidth}}
	\caption{Performances of the proposed SARAD and DQN on three indicators: average running time per episode, average reward per step, and total reward per episode.}
	\label{fig:4}
\end{figure*}

\subsection{Main Results}
To demonstrate the superiority of the proposed framework and the contributions of individual components, the proposed models are compared against vanilla DQN and performed ablation studies across different configurations, with detailed model setups summarized in Table \ref{tab:1}.
\begin{table}[htbp]
  \centering
  \caption{\normalfont Model configurations for comparison.}
    \begin{tabular}{llll}
    \toprule
    Discrimination & Collision Predictor & RAG Strategy & Model Name \\
    \midrule
    \XSolidBrush  & \XSolidBrush  & \XSolidBrush  & Vanilla DQN \\
    \XSolidBrush  & GBDT+LR & \XSolidBrush  & SARAD-G \\
    \checkmark     & GBDT+LR & \XSolidBrush  & SARAD-DG \\
    \checkmark     & GBDT+LR & HNSW+MMR & SARAD-DGH \\
    \checkmark     & LLM (Qwen0.6B) & HNSW+MMR & SARAD-DLH \\
    \bottomrule
    \end{tabular}%
    \label{tab:1}%
\end{table}%

Performance comparisons between the SARAD variants and vanilla DQN are visualized in Fig. \ref{fig:4}. Note that to clearly illustrate the specific learning dynamics and algorithmic behaviors without visual clutter, the learning curves represent a single representative training run for each configuration rather than an average across multiple random seeds. The results indicate that all algorithms exhibit significant upward trends in both running time and rewards throughout the training, confirming effective policy learning. However, notable disparities emerge in convergence speed, stability, and final performance. Specifically, Vanilla DQN exhibits the poorest initial learning performance and suboptimal early-stage progress. Although achieving stable improvement and eventual convergence, its overall performance remains slightly inferior to SARAD series. In contrast, SARAD-DLH, equipped with an LLM-based Collision Predictor, demonstrates the best overall performance, maintaining higher metrics with exceptional stability and converging to superior values throughout training. SARAD-DLH achieves the highest average runtime and competitive reward performance during late-stage training. Notably, it converges faster to a stable policy in terms of average runtime of each episode, while exhibiting a more conservative learning curve on reward-related metrics during the initial training phase. SARAD-DGH and SARAD-DG rank second in overall efficacy, whereas the vanilla DQN exhibits reliable but comparatively slower learning capabilities.
\begin{table}[htbp]
  \centering
  \caption{\normalfont Performances of the whole process.}
    \begin{tabular}{lccc}
    \toprule
    \multicolumn{1}{c}{Model Name} & \multicolumn{1}{p{5.5em}}{Avg Length} & \multicolumn{1}{p{5.8em}}{Avg Reward} & \multicolumn{1}{p{6.8em}}{Total Reward} \\
    \midrule
    Vanilla DQN & 21.5657  & 0.9203  & 20.6767  \\
    SARAD-G & 22.0195  & 0.9163  & 20.8875  \\
    SARAD-DG & 22.5377  & 0.9183  & 21.3811  \\
    SARAD-DGH & 22.5228  & \textbf{ 0.9229 } & \textbf{ 21.5301 } \\
    SARAD-DLH & \textbf{ 22.7478 } & 0.9132  & 21.4075  \\
    \bottomrule
    \end{tabular}%
    \label{tab:2}%
\end{table}%

Quantitative results for the entire training process are presented in Table \ref{tab:2}, where SARAD-DGH attains the highest reward metrics, while SARAD-DLH excels in average running time. To further evaluate the advantages of the proposed models in the late-training, we specifically assessed performance over the last 80,000 steps, with detailed metrics provided in Table \ref{tab:3}. 
\begin{table}[htbp]
  \centering
  \caption{\normalfont Performances for the last 80k steps.}
    \begin{tabular}{lccc}
    \toprule
    \multicolumn{1}{c}{Model Name} & \multicolumn{1}{p{5.5em}}{Avg Length} & \multicolumn{1}{p{5.8em}}{Avg Reward} & \multicolumn{1}{p{6.8em}}{Total Reward} \\
    \midrule
    Vanilla DQN & 23.2547  & 0.9417  & 22.5967  \\
    SARAD-G & 24.3784  & \textbf{ 0.9501 } & 23.8556  \\
    SARAD-DG & 24.4080  & 0.9475  & 23.8098  \\
    SARAD-DGH & 24.3857  & 0.9485  & 23.8439  \\
    SARAD-DLH & \textbf{ 24.8768 } & 0.9479  & \textbf{ 24.1935 } \\
    \bottomrule
    \end{tabular}%
    \label{tab:3}%
\end{table}%

It can be seen that SARAD-DLH, leveraging an LLM as the Collision Predictor, delivers the most robust comprehensive performance during advanced training stages. Although LLM invocations introduce inference latency, the $\epsilon$-expert exploration strategy confines this computational overhead primarily to the early exploration phase, ensuring real-time feasibility for the converged policy.

\section{Conclusion}
This paper proposes SARAD, a novel framework that leverages LLMs to enhance DRL training. By integrating retrieval-augmented generation for LLM-based decision-making with an attention discriminator-guided DRL policy network, the framework mitigates uncertainties caused by early-stage random exploration. It also accelerates the stabilization of average runtime per episode and improves overall learning efficacy. To enhance robustness, the design incorporates a reflective module and a collision predictor, collectively maximizing operational safety for autonomous vehicles. Evaluations on the highway-env environment demonstrate the framework's compelling performance, with future work focused on developing more efficient inference architectures.

\bibliographystyle{IEEEtran}
\bibliography{ref-kanzie}
\end{document}